\documentclass[11pt]{article}
\usepackage{amscd,amssymb,stmaryrd}

\usepackage{graphicx}
\usepackage{subcaption}
\usepackage[utf8]{inputenc}
\usepackage[export]{adjustbox}
\usepackage{wrapfig}

\usepackage{amsthm}

\usepackage{color}
\usepackage{mathtools}

\usepackage{hyperref}
\usepackage[english]{babel}
\usepackage{booktabs}

\usepackage{float}
\usepackage[linesnumbered,ruled,vlined, ruled,vlined]{algorithm2e}

\theoremstyle{definition}

\usepackage{tikz}
\usepackage{pifont}%

\usepackage{diagbox}
\usetikzlibrary{decorations.pathreplacing,calligraphy}
\usepackage{pgfplots}
\pgfplotsset{compat=1.11}
\usepgfplotslibrary{groupplots}

\usetikzlibrary{shapes.misc}

\tikzset{cross/.style={cross out, draw=black, minimum size=2*(#1-\pgflinewidth), inner sep=0pt, outer sep=0pt},
cross/.default={1pt}}

\title{MODNO: Multi-Operator Learning With Distributed Neural Operators}

\author{Zecheng Zhang\footnote{Department of Mathematics, Florida State University, Tallahassee, FL 32304, USA. (Email: zecheng.zhang.math@gmail.com)} }

\begin{document}

\maketitle

\begin{abstract}

The study of operator learning involves the utilization of neural networks to approximate operators. Traditionally, the focus has been on single-operator learning (SOL). However, recent advances have rapidly expanded this to include the approximation of multiple operators using foundation models equipped with millions or billions of trainable parameters, leading to the research of multi-operator learning (MOL).
In this paper, we present a novel distributed training approach aimed at enabling a single neural operator with significantly fewer parameters to effectively tackle multi-operator learning challenges, all without incurring additional average costs. Our method is applicable to various neural operators, such as the Deep Operator Neural Networks (DON).
The core idea is to independently learn the output basis functions for each operator using its dedicated data, while simultaneously centralizing the learning of the input function encoding shared by all operators using the entire dataset.
Through a systematic study of five numerical examples, we compare the accuracy and cost of training a single neural operator for each operator independently versus training a MOL model using our proposed method. Our results demonstrate enhanced efficiency and satisfactory accuracy. Moreover, our approach illustrates that some operators with limited data can be more effectively constructed with the aid of data from analogous operators through MOL learning. This highlights another MOL's potential to bolster operator learning.
\end{abstract}

\section{Introduction}
Operator learning \cite{chen1995universal, chen1993approximations, lu2021learning, li2020fourier, zhang2023belnet}, or more specifically, single-operator learning (SOL), is a new area in scientific machine learning \cite{raissi2019physics, schaeffer2013sparse, schaeffer2017learning, schaeffer2017sparse, leung2022nh, efendiev2022efficient, zhang2019convergence} that learns the operator mapping from one function to another function. 
The early study in the area is the universal approximation theorem for functionals \cite{chen1993approximations} and the operator proposed in \cite{chen1995universal}. 

Operator learning has many applications in solving classic challenging mathematical and scientific machine learning (MSML) problems. One of the most addressed problems is to solve the parametric PDE problem \cite{li2020neural, bhattacharya2021model, lanthaler2023curse, kovachki2021universal, de2023convergence, liu2024deep, hasani2024generating}. For example, researchers want to learn the mapping from the parametrized initial condition to its corresponding solutions \cite{wang2021learning, li2021physics, lu2022comprehensive, li2020fourier}. Researchers also successfully applied the neural operators to solve large-scale real-life applications such as weather \cite{pathak2022fourcastnet}, geology \cite{zhu2023fourier, jiang2023fourier, li2023solving}, physics \cite{mao2023ppdonet, mao2021deepm, di2023neural}, material science \cite{lee2023deep, lu2023deep}, power systems and control \cite{lin2023b, moya2023approximating, chen2023neural} etc.
Besides the application, the study of operators has extended to many other areas. For example, the approximation with the neural operator \cite{deng2022approximation,lanthaler2022error, zhang2023discretization, lanthaler2022nonlinear, de2023convergence,bhattacharya2021model, liu2024deep, lanthaler2024neural}; the uncertainty quantification and optimization \cite{lin2023b, moya2024conformalized, ma2024calibrated, garg2023vb, lin2022multi, chen2023operator}, etc.

Over recent years, numerous neural operator models have emerged, such as the Deep Operator Neural Network (DON) and its various iterations \cite{lu2021learning, lu2022comprehensive, jin2022mionet, zhang2023belnet, zhang2023discretization, zhu2023reliable, howard2022multifidelity, li2022learning}, as well as the Fourier Neural Operator and its derivatives \cite{li2020fourier, li2020neural, wen2022u, li2022fourier, pathak2022fourcastnet, lin2023b, zhang2023d2no, li2021physics, li2023fourier, li2023solving}. Please refer to \cite{qiu2024derivative, patel2024variationally, o2022derivative, hua2023basis, tripura2023wavelet} for some other notable structures.
DNO stands out for not requiring a discretization for the output functions \cite{lu2021learning, zhang2023belnet}, thereby granting researchers greater adaptability in tackling scientific computing challenges. For instance, a trained DON could serve as a mesh-free numerical solver for PDEs \cite{zhang2023homogenization, lu2022multifidelity, howard2022multifidelity}. 
The vanilla DON \cite{lu2021learning} lacked discretization invariance, necessitating uniform discretizations across all input functions. However, recent advancements \cite{zhang2023belnet, zhang2023discretization} in universal approximation \cite{chen1995universal} and distributed training algorithms \cite{zhang2023d2no} have rendered DON discretization-invariant.

Multi-operator learning (MOL) \cite{liu2023prose, yang2023prompting, mccabe2023multiple} represents a broader extension of SOL, allowing the simultaneous learning of multiple operators. MOL has received tremendous attention owing to its potential to generalize and extrapolate to new operators without further training \cite{zhu2023reliable}. Notably, PROSE \cite{liu2023prose} has demonstrated the ability to forecast operators resembling those within the training set and extend predictions to operators whose output functions' physical properties are not encountered during training.

Because of the encoding complexity of operators, the majority of MOL models possess millions or even billions of trainable parameters and necessitate high-performance computational devices for training. In this study, we introduce a distributed training technique termed ``Multi-Operator Distributed Neural Operator (MODNO)'', aimed at empowering standard single neural operators with significantly fewer parameters \cite{lu2021learning, jin2022mionet, zhang2023belnet, zhang2023discretization} to address multi-operator learning challenges.

Distributed learning \cite{mcmahan2017communication, verbraeken2020survey,yin2021comprehensive, moya2022fed}, also referred to as distributed training, is a paradigm in machine learning where the training process is distributed across multiple computing nodes or devices. Unlike traditional centralized training where all computations are performed on a single machine, distributed learning leverages the computational power of multiple devices to accelerate the training process and handle large-scale datasets efficiently.

Researchers have adopted the distributed training idea in solving scientific computing problems, particularly the operator learning problem. For example, in \cite{zhang2023d2no}, the authors propose to employ different neural networks to handle heterogeneous input functions with distinct properties, while using a centralized neural network to learn the common shared basis for output functions corresponding to all input functions. This method enables the DON to be discretization invariant.

In this study, we propose to employ a distributed approach to learn multiple operators within a single DON framework. 
The concept involves training specialized output basis functions for each operator using dedicated data associated with that operator, while simultaneously employing a shared neural network to encode the input functions.
One crucial analysis often overlooked in MOL studies is the comparison with training separate SOL neural operators using distinct datasets. We demonstrate that MODNO may outperform this approach in many cases (11 out of 16) even when provided with less data. We summarize the contributions as follows.

\begin{itemize}
\item We propose a distributed training approach named ``Multi-operator learning with distributed neural operators (MODNO)'' to learn multiple operators utilizing a single neural operator (SNO). This training method is applicable to all Chen-Chen-type neural operators and has significantly less cost when compared to many other multi-operator learning (MOL) foundation models.

\item We compute the training cost, such as the total number of back-propagations and forward passes in gradient flow and trainable parameters, for constructing one operator within the MODNO framework. Our analysis reveals that, despite requiring less cost compared to training an SNO independently, the framework achieves similar or even superior accuracy.

\item Through the numerical experiments, we illustrate that MOL may boost the accuracy for specific operators compared to training these operators separately using individual SNOs. This finding indicates a new advantage of MOL, suggesting its potential to enhance predictions for certain operators by leveraging data from other operators.
\end{itemize}

The rest of the paper is organized as follows. In Section \ref{sec_review}, we will review the DON structure and introduce the motivation of the algorithms based on the approximation theorem \cite{chen1995universal}. Later in Section \ref{sec_methodology}, we introduce the algorithm and discuss the computation costs of the algorithms. We will then present the numerical results in Section \ref{sec_numerical}. Finally, we will discuss limitations and future works in Section \ref{sec_conclusion}.

\section{Overview}
\label{sec_review}

\subsection{(Single) Operator Learning (SOL)}
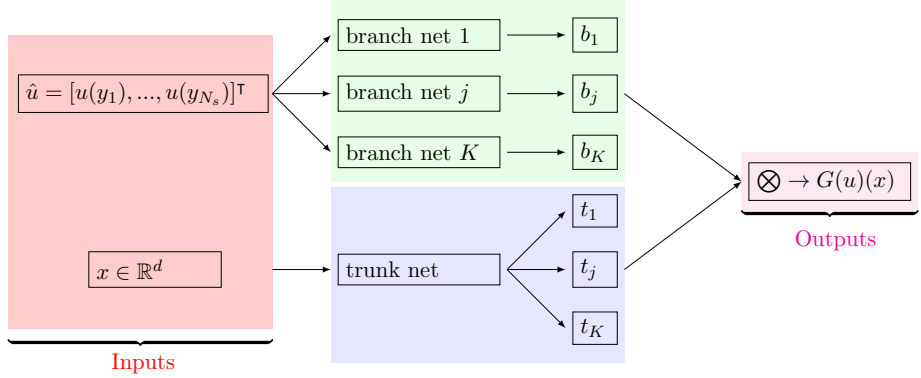
\begin{figure}[H]
\centering
\scalebox{.78}{\begin{tikzpicture}[scale = 1]
 \fill [green!10] (2, 4.5) rectangle (7, 7.6);
  \fill [blue!10] (2, 1.4) rectangle (7, 4.4);
    \fill [red!20] (-3.5, 2) rectangle (1, 7);

    \draw[ultra thick] [decorate,
    decoration = {calligraphic brace, mirror}] (-3.5, 1.8) --  (1, 1.8);
\node at (-1.2, 1.4) {\textcolor{red}{Inputs}};

\fill [magenta!10] (9, 4) rectangle (12, 5);
    \draw[ultra thick] [decorate,
    decoration = {calligraphic brace, mirror}] (9, 4) --  (12, 4);
\node at (10.6, 3.5) {\textcolor{magenta}{Outputs}};

\node[draw, text width=4cm] at (-1.2, 6) {$\hat{u} = [u(y_1), ..., u(y_{N_s})]^\intercal$};

 \draw [-latex ](1,6) -- (2, 7);
 \node[draw, text width = 2.5cm] at (3.5, 7) {branch net $1$};
 \draw [-latex ](5, 7) -- (6, 7);
 \node[draw, text width = 0.5cm] at (6.5, 7) {$b_1$};
 
 \draw [-latex ](1,6) -- (2, 5);
  \node[draw, text width = 2.5cm] at (3.5, 5) {branch net $K$};
  \draw [-latex ](5, 5) -- (6, 5);
  \node[draw, text width = 0.5cm] at (6.5, 5) {$b_K$};
  
  \draw [-latex ](1,6) -- (2, 6);
 \node[draw, text width=2.5cm] at (3.5, 6) {branch net $j$};
 \draw [-latex ](5, 6) -- (6, 6);
\node[draw, text width = 0.5cm] at (6.5, 6) {$b_j$};

 \node[draw, text width = 2cm] at (-1, 3) {$x\in\mathbb{R}^d$};
  \draw [-latex ](1, 3) -- (2, 3);
 \node[draw, text width = 2.5cm] at (3.5, 3) {trunk net};
 
 \draw [-latex ](5, 3) -- (6, 4);
\node[draw, text width = 0.5cm] at (6.5, 4) {$t_1$};

  \draw [-latex ](5, 3) -- (6, 3);
\node[draw, text width = 0.5cm] at (6.5, 3) {$t_j$};

   \draw [-latex ](5, 3) -- (6, 2);
 \node[draw, text width = 0.5cm] at (6.5, 2) {$t_K$};

\draw [-latex ](7, 6) -- (9, 4.5);
  \draw [-latex ](7, 3) -- (9, 4.5);

  \node[draw, text width = 2.5cm] at (10.5, 4.5) {$\bigotimes\rightarrow G(u)(x)$};

\end{tikzpicture}}
\caption{Stacked version DON \cite{lu2021learning}, a fundamental work in the area. $\bigotimes$ denotes the inner product in $\mathbb{R}^K$.}
\label{fig_don_structure}
\end{figure}

The goal of single-operator learning (SOL) is to approximate the operator  $G: U\rightarrow V$. Here $U$ and $V$ are function spaces with domain $D$ and $D'$ separately.
In \cite{chen1995universal, lu2021learning}, the authors Chen and Chen proposed the following approximation to the operator:
\begin{align}
    G(u)(x) \approx \sum_{k = 1}^K a_k(u)b_k(x),
    \label{eqn_approx}
\end{align}
where $u\in U$ and $x\in D'$, and $a_k(\cdot)$ and $b_k(\cdot)$ are functionals on $U$ and functions on $D'$ respectively with some regularity assumptions (please refer to Theorem 5 in \cite{chen1995universal}, Theorem 3.8 in \cite{zhang2023discretization}, Theorem 2 of \cite{lu2021learning}, and Theorem 2.10 in \cite{jin2022mionet}).
By employing neural networks to approximate $a_k(\cdot)$ and $b_k(\cdot)$, various network architectures have been developed by researchers \cite{lu2021learning, lu2022comprehensive, zhang2023belnet, zhang2023discretization, jin2022mionet, wang2021learning}. Among these, the fundamental work is the Deep Operator Neural Network (DON) \cite{lu2021learning, lu2022comprehensive, jin2022mionet, zhu2023reliable} and its variants. Since all mentioned structures share the approximation structure, we will call them Chen-Chen-type neural operators, and denote the network as $G_\theta$, with $\theta$ being the parameters, specifically, 
\begin{align*}
     G(u)(x) \approx G_{\theta}(\hat{u})(x) = \sum_{k = 1}^K a_k(\hat{u}; \alpha)b_k(x; \beta).
     \label{eqn_network_approx}
\end{align*}
Here $\hat{u} = [u(y_1), ..., u(y_{N_s})]^{\intercal}$, and $\{y_i\}_{i = 1}^{N_s}\subset D$ are sensors.
$a_k(\cdot; \alpha)$ and $b_k(\cdot; \beta)$ are neural network parametrized functions with parameter $\alpha$ and $\beta$ that approximate $a_k$ and $b_k$ in Equation (\ref{eqn_approx}). 
In this study, we will specifically employ the DON structure \cite{lu2021learning}, which has been demonstrated to be highly successful across a range of applications and theoretical frameworks. In the DON literature, $a_k(\cdot; \alpha)$ is called the branch net and $b_k(\cdot; \beta)$ is called the trunk net. Specifically,
let us denote $x\in\mathbb{R}^d\subset D'$, $g$ be the Tauber-Wiener function \cite{chen1995universal}. It follows that,
\begin{align}
    b_k(x) & = g(\omega_k\cdot x+ \zeta_k),\\
    a_k(\hat{u}) &= \sum_{l = 1}^M c_l^k\, g\left(\sum_{j = 1}^{N_s}\varepsilon_{lj}^ku(y_j)+\theta_l^k\right),
    \label{eqn_br_approx}
\end{align}
here $c_l^k, \zeta_k, \theta_l^k, \varepsilon_{lj}^k\in\mathbb{R}$, $\omega_k\in\mathbb{R}^d$ are parameters. We present one picture demonstration of DON in Figure \ref{fig_don_structure}, and will design a novel distributed learning algorithm to extend the DON to be a multi-operator learning framework with significantly less trainable parameters. 
Notably, this method can be generalized to all Chen-Chen-type neural operators such as the Basis Enhanced Learning (Bel) \cite{zhang2023belnet, zhang2023discretization} the discretization invariant extension of DON.

\subsection{Multi-Operator Learning (MOL)}
The multi-operator learning \cite{liu2023prose, yang2023context,yang2023prompting,yang2024pde, ye2024pdeformer, shen2024ups, mccabe2023multiple} is a further generalization of SOL. MOL constructs a single neural network and learns multiple operators simultaneously. 
Specifically, suppose we have $N_{op}$ operators and $i-th$ operator is defined as $G_i: U_i\rightarrow V_i$, where $U_i$ and $V_i$ are input and output function spaces of $G_i$. The goal of MOL is to design a neural network $\mathcal{N}$ that takes $G_i$ and $u_{ij}\in U_i$ as inputs and constructs the output function $G_i(u_{ij})\in v_i$. Due to the encoding of the operator $G_i$, the current MOLs models are usually built with millions even billions of parameters, and trained with huge numbers of samples. For example, the ICON \cite{yang2023context} has around $30$ millions parameters, and PROSE-PDE \cite{liu2023prose}
has around $105$ millions parameters.

The development of large MOL models has been contributing numerous innovative concepts and techniques to the MSML community. For instance, the adoption of symbolic encoding of operators proposed in \cite{liu2023prose, song2024finite}, the advocacy for using graphs to encode operators as highlighted in \cite{ye2024pdeformer}, and fine-tuning of the foundation large language models (LLMs) \cite{yang2023prompting, yang2023context}.
More importantly, MOL has received a lot of attention as it has the potential to extrapolate to unseen operators. Some MOL models for example PROSE \cite{liu2023prose} have shown that PROSE can predict operators unseen but similar to the training dataset operators. 

However, most MOLs still struggle to handle challenging extrapolation tasks. Consequently, it is advocated and important to compare the training costs (for example the number of parameters) and prediction accuracy of employing a single large foundation MOL for multiple operators versus training individual SOLs for each operator. In this study, we will devise a training method to empower a single Chen-Chen-type neural operator to address the challenges of multi-operator learning and subsequently evaluate its performance. Notably, compared to the well-known MOL foundation models, the proposed framework has significantly fewer trainable parameters and may even outperform individually trained DON for each single operator.

\section{Methodology}
\label{sec_methodology}
In this section, we will outline the proposed approach and present the metric to evaluate the cost of the algorithms. 

\subsection{Theoretical Motivation}
\label{sec_motivation}
As discussed in Section \ref{sec_review}, Equation (\ref{eqn_approx}) demonstrates that a single operator can be effectively approximated. This involves crafting distinct networks tailored to approximate $a_k(\cdot)$ and $b_k(\cdot)$, thereby creating different neural operators. 

To establish the universal approximation theory of \cite{chen1995universal, zhang2023discretization, jin2022mionet},
the final step is to approximate the function $G(u)\in V$ using neural networks, as 
$G(u) = \sum_{k = 1}^K c_k\circ G(u) b_k(x)$, where $u\in U$,  and $c_k(\cdot)$ is learnable functional on $V$. 
Notably, $b_k(\cdot)$ or trunk nets serve as the learnable basis for constructing the output function in $V$, and is independent of the input functions $u\in U$, i,e., all $G(u)$ share the basis.

For the functional $c_k\circ G(u)$, by assuming a finite-dimensional structure $\hat{u}\in \mathbb{R}^{N_s}$ of $u$, the functional $c_k\circ G(u)$ can be approximated by the branch nets as in Equation (\ref{eqn_br_approx}), or, 
\begin{align*}
    a_k(\hat{u}) = \sum_{l = 1}^M c_l^k\, g\left(\sum_{j = 1}^{N_s}\varepsilon_{lj}^ku(y_j)+\theta_l^k\right) := \sum_{l = 1}^Mc_l^ke_l^k(u),
\end{align*}
where $e_l^k(u)$ can be regarded as the basis for an encoding of the input function space $U$. 

Motivated by the above approximation theory,
with multiple operators $G_i$, one should design dedicated unshared trunk networks to approximate the basis $b_k(\cdot)$ for each operator's output functions, utilizing the associated data for each operator. 
Assuming all input functions share the input function space, one can use centralized networks to learn a shared centralized input function basis $e_l^k(u)$ that is applicable across all distinct operators. 

In the implementation phase, an alternative strategy involves federately learning all branch networks $a_k(\cdot;\alpha)$ rather than concentrating on the input function space basis $e_l^k(\cdot)$, potentially leading to additional computational savings. Further elaboration on the cost savings will be provided in Section \ref{sec_algo1}.

\subsection{Multi-Operator Learning Distributed Neural Operator (MODNO)}
\label{sec_algo1}
The approximation to each individual operator $G_i: U_i\rightarrow V_i$ is the same as Equation \ref{eqn_network_approx}, but the trunk basis and part of branch nets will be dedicated to each operator.
To introduce the algorithm
let us first introduce the local loss functions associated with operator $G_i$.

The local loss is used to train unshared networks to approximate the $k-$th basis $b_{k, i}(\cdot)$ for $i-th$ operator, and let us denote the network approximation to $k-$th basis $b_{k, i}(\cdot)$ as $b_{k,i}(\cdot, \beta_{k,i})$, where $\beta_{k,i}$ denotes the trainable parameters, and $\beta_i = \cup_k \beta_{k, i}$.
Additionally, we use $a_{k}(\cdot; \alpha_k)$ to denote the centralized shared network used to approximate $a_k(\cdot)$, and trainable parameters denote as $\alpha = \cup_k \alpha_k$.
It is worth noting that in this context, we make all parameters in the branch networks shared by all, contrasting the motivation outlined in Section \ref{sec_motivation}, yet proving to be more efficient. More details are presented in Section \ref{sec_cost}.

Additionally, for demonstration and without loss of generality, we use only one point $x_i\in D'_i$ for the $i-th$ operator, where $D'_i$ is the domain for functions in $V_i$. We then can define the local loss with suitable normalizing as follows,
\begin{align*}
    L_i(\beta_i; \alpha) = \sum_{p = 1}^{N_u}\|G_i(\hat{u}_{i, p})(x_i) - \sum_{k = 1}^K a_k(\hat{u}_{i, p}; \alpha_k) b_{k,i}(x_i; \beta_{k, i}) \|^2,
\end{align*}
here $N_u$ is the total number of input functions and $\hat{u}_p$ denotes discretization of the corresponding functions $u_{i, p}\in U_i$. This local loss is used to update the dedicated parameters $\beta_i$ given the shared parameters $\alpha$, and the updating is based on the data associated with $i-th$ operator only.

Now, let us define the global loss function $L$, we use the global loss to train the centralized shared network $a_k(\cdot; \alpha_k)$,
\begin{align*}
    L(\alpha; \beta) = \sum_{i = 1}^{N_{op}} L_i(\beta_i; \alpha),
\end{align*}
here $N_{op}$ is the number of operators. 
The global loss is used to update the globally shared parameters given all dedicated parameters $\beta_i$, with the update process relying on all data associated with all operators. 
Let us now summarize the pseudo-algorithm with standard gradient descent optimization in Algorithm \ref{algo_unshared}. Extending it to other gradient descent-based algorithms is straightforward.

\begin{algorithm}[H]
\caption{Multi-Operator Learning Distributed Neural Operator (MODNO)}
\label{algo_unshared}
Initialization: parameters $\beta_i$ for $i-th$ operator's dedicated network (e.g., trunk nets), $\alpha$ for the shared network (e.g., branch nets), the optimization epochs $N_I$, learning rate $\eta_i$ and $\eta$\;
\For{$epoch = 1$ to $N_I$}
{
    Update the unshared parameters $\beta_i$ for operator $G_i$ using its own dataset as follow\;

\For{$i = 1$ to $N_{op}$}
{
$$
\beta_i = \beta_i - \eta_i \nabla_{\beta_i} L_i(\beta_i;\alpha).
$$
}

Update the shared networks parameters $\alpha$ with all data as follow\;
$$
\alpha = \alpha  - \eta\nabla_{\alpha} L(\alpha; \beta).
$$
}

\end{algorithm}

\subsection{Computational Cost}
\label{sec_cost}
When calculating the costs of training two distinct neural networks, the conventional practice involves quantifying the total number of trainable parameters and training samples for each network. Nonetheless, within the context of the proposed distributed algorithm, a straightforward calculation of total parameters across all shared and dedicated networks proves inadequate, mainly because the dedicated networks undergo training with a reduced number of training samples (their dedicated dataset).

To ensure a fair assessment of the distributed algorithm's efficacy, we instead prioritize the number of back-propagations and forward passes \cite{higham2019deep} as the metric for the cost and comparison with a single DON.

To evaluate the total number of back-propagations and forward passes for MODNO, we denote the following variables for the datasets and training samples: $N_{op}$ for the number of operators, $N_{u, i}$ for the number of functions for $i-$th operator, and $N_{p, i, j}$ for the number of known training function values of the $j$-th output function of the $i$-th operator.

To present a more general concept, with a slight abuse of notation without introducing ambiguity, $b_i(\cdot;\beta_i)$ denotes the dedicated networks, but are not necessarily limited to the trunk net in the previous sections, and we use $N_{b, i}$ to denote the number of back-propagations and forward passes for one training sample for $b_i(\cdot;\beta_i)$. Similarly, $N_{a}$ denotes the number of back-propagations and forward passes for one sample for all shared nets $a(\cdot; \alpha)$.

We will compare the total number of back-propagations and forward passes for training $N_{op}$ individual single-operator learning DONs and one for each operator. 
We then denote $N_{u, i}'$ for the number of functions for $i-$th SOL DON, and $N_{p, i, j}'$ for the number of function values of the $j$-th output function of the $i$-th SOL DON.
Besides, the back-propagation and forward pass counts for $i-$th individual DON's branch nets and trunk nets are $N_{b, i}'$ and $N_{a, i}'$.

The cost for MODNO is then $C_{MOL} = \sum_{i = 1}^{N_{op}} N_{u, i}N_{u, i, j} (N_{b, i}+ N_a)$, and total cost of training $N_{op}$ individual DONs for each operator is $C_{SOL} = \sum_{i = 1}^{N_{op}} N_{u, i}'N_{u, i, j}' (N_{b, i}'+ N_{a, i}')$. Assuming MODNO shares the identical structure with individual DONS and shares the training samples, $C_{MOL} = C_{SOL}$. 
However, we observe from the numerical experiments that MODNO generally achieves better accuracy and remains accurate even if we reduce $N_{u, i}<N_{u, i}'$ and $N_{u, i, j}<N_{u, i, j}'$.

In particular, given that MODNO and all individual DONs have the same structure, MODNO's centralized shared network $\alpha$ is trained with a significantly larger dataset compared to single DONs. Since the single DON can achieve satisfying results for many tasks, together with the overfitting concerns for $\alpha$, we propose to reduce the number of training samples for the centralized shared structure. Our numerical experiments indicate that despite reducing the training samples, prediction errors remain robust; however, fewer samples contribute to improved efficiency. Specifically, the new cost for MODNO is: $C_{MOL} = \sum_{i = 1}^{N_{op}} N_{u, i}N_{u, i, j}N_{b, i}+ qN_{u, i}N_{u, i, j}N_a$, here $q\in[0, 1]$ is the percentage of the data used in training the shared structure.
In summary, this approach motivates us to share more networks (i.e., make $N_a$ larger) and utilize less data for the centralized structure. Consequently, we make the entire branch networks shareable, and it has proved to be effective in numerical experiments.

\section{Numerical Experiments}
\label{sec_numerical}

In this section, we present numerical evidence to validate our algorithm. We will analyze five examples, with the first four being experiments involving learning three operators. The final experiment (the most challenging one) focuses on learning four operators, and we also test the extrapolation to predict the solution at a time out of the training time range. 
Additionally, the testing spatial mesh is different from the training solution spatial mesh in all experiments.
For each example, we will compare MODNO's performance to training separate DONs for each operator, all having the same architecture. To ensure reproducibility, all code and experiment results will be made available on Google Colab upon publication.

\subsection{Example One}
\label{sec_wave_klein_sine}
\begin{figure}[H]
    \centering
    \includegraphics[scale = 0.4]{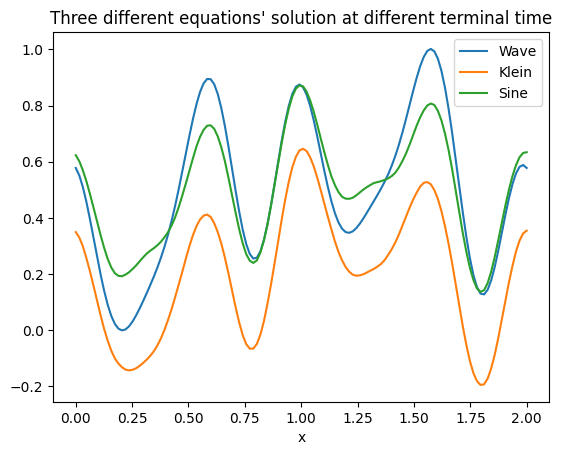}
    \caption{The terminal solutions of the Wave equation (\ref{eqn_wave}), Klein Gordon equation (\ref{eqn_klein}) and Sine-Gordon equation (\ref{eqn_sine}) with one same initial condition.  }
    \label{fig_wave_klein_sine}
\end{figure}

In this section, we will study the following three equations and the operator of mapping the initial condition to the solutions at a later time. The wave equation,
\begin{align}
    u_{tt}  = u_{xx}, t\in [0, 1], x\in [0, 2],
    \label{eqn_wave}
\end{align}
the Klein-Gordon equation, $m=0.1$, $c = 10$,
\begin{align}
    u_{tt} + m^2c^4u = c^2u_{xx}, t\in [0, 2], x\in [0, 2],
    \label{eqn_klein}
\end{align}
and the Sine-Gordon equation,
\begin{align}
    u_{tt} + c\sin(u) = u_{xx}, t\in[0, 2], x\in [0, 2].
    \label{eqn_sine}
\end{align}
All equations are equipped with the periodic boundary conditions.
The initial conditions are generated using the Fourier series type function. Specifically, 
$u_0(x; w_i) = w_1\sin(\pi x)+w_2\sin(2\pi x)+w_3\sin(4\pi x)+w_4\sin(6\pi x)+ w_5\cos(\pi x)+w_6\cos(2\pi x)+w_7\cos(4\pi x)+w_8\cos(6\pi x)+w_9$, and $w_i\sim\mathcal{U}(-2, 2)$, $i = 1,..., 8$ and $w_9\sim\mathcal{U}(0.1, 2)$. We present one realization of the solution of three different equations at the terminal time in Figure \ref{fig_wave_klein_sine}.
To measure the variation of three equations' solution data, we also compute the mean of all training solutions and use it as a prediction for all three different equation solutions. The mean relative error for Wave, Klein, and Sine are $49.78\%$, $99.85\%$, and $40.37\%$ respectively. We finally present the numerical results in Table \ref{table_wave_klein_sine}.

\begin{table}[H]
    \centering
    \begin{tabular}{|c|p{2.cm}|p{2.3cm}|p{2.3cm}|p{2.3cm}|p{2.3cm}|}
    \hline
     Operator & Single DON $100\%$ data & MODNO $100\%$ data &  MODNO $90\%$ data &  MODNO $80\%$ data &  MODNO $70\%$ data \\
    \hline
    Wave (\ref{eqn_wave}) & $2.59\%$ & $2.23\%$ & $2.26\%$ & $2.44\%$ & $2.56\%$ \\
    \hline
    Klein (\ref{eqn_klein}) & $4.12\%$ & $3.69\%$ & $4.03\%$ & $4.0\%$ & $4.12\%$ \\
    \hline
    Sine (\ref{eqn_sine}) & $2.35\%$ & $2.09\%$ & $2.02\%$& $2.78\%$ & $1.92\%$ \\
    \hline
    \end{tabular}
    \caption{Results for experiment one. The second column, Single DON, is to train three individual DONs with identical structures with full data for each operator. Notably, when reducing MODNO data (i.e., $q = 0.9, 0.8, 0.7$), MODNO has lower costs when compared to three individual DONs. Please refer to Section \ref{sec_cost} for the cost evaluation details.} 
    \label{table_wave_klein_sine}
\end{table}

\subsubsection{Experiment One Results Analysis}
We compare training one multi-operator MODNO with training three separate DONs one for each operator with identical network structure.
We can observe from Table \ref{table_wave_klein_sine} that MONDNO outperforms training three individual DONs given the same amount of data. Specifically, $2.23\%$ vs $2.59\%$ for Wave, $3.69\%$ vs $4.12\%$ for Klein-Gordon equation, and $2.09\%$ vs $2.35\%$ for Sine Gordon equation. 

When we decrease the data utilized in training the shared structure, the prediction accuracy for the Sine Gordon equation remains resilient to the data reduction, with only a slight increase in errors observed for the Wave and Klein Gordon equations. 
Nonetheless, a significant advantage stemming from the enhanced efficiency with reduced data usage is evident. Even with a dataset approximately $80\%$ smaller than that employed for single DONs in multi-operator learning, MODNO exhibits superior accuracy in predicting solutions for the Wave and Klein equations.

We do not observe this improved accuracy and enhanced efficiency in every single operator,
but this trend is consistently observed for most of the numerical experiments we conducted, highlighting a potential advantage and novelty of multi-operator learning within the MOL frameworks. Specifically, multi-operator learning exhibits the capability to predict solutions with satisfactory accuracy even when provided with less data compared to single-operator learning approaches.

\subsection{Experiment Two}
\begin{figure}[H]
    \centering
    \includegraphics[scale = 0.4]{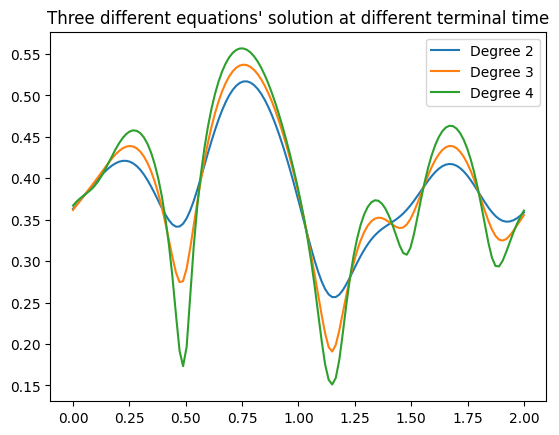}
    \caption{The terminal solutions of three porous media equations (\ref{eqn_porous_media}) with different degrees (two, three, and four) with one same initial condition.  }
    \label{fig_porous_media}
\end{figure}

In this section, we will examine another three nonlinear time-dependent equations along with the operator that maps the initial condition to the solution at a later time; to test the extrapolation, the testing solution time is beyond the training time range. Specifically, we will delve into the porous media type equation with different order $m = 2,3, 4$,
\begin{align}
    u_t = (u^m)_{xx}, x\in [0, 2], t\in [0, 0.01].
    \label{eqn_porous_media}
\end{align}
All equations are equipped with periodic boundary conditions.
We generate the initial condition using the Fourier series type functions by uniformly sampling the parameters in $[-1, 1]$. Specifically, $u(x; w_i) = 0.1(w_1\sin(\pi x)+w_2\sin(2\pi x)+w_3\sin(3\pi x)+ w_4\cos(2\pi x)+ w_4\cos(4\pi x)+w_6\cos(6\pi x))$, and $w_i\sim \mathcal{U}(-1, 1)$. We present one realization of all three equations' solutions in Figure \ref{fig_porous_media}.

We first calculate the mean of all training target solutions, and use this solution to predict the testing solutions of three equations separately. The relative $L_2$ errors for degree two, three and four porous media equations are $15.43\%$, $17.93\%$ and $20.29\%$ respectively. We then present the numerical results in Table \ref{table_porous_media}.


\begin{table}[H]
    \centering
    \begin{tabular}{|c|p{2.cm}|p{2.3cm}|p{2.3cm}|p{2.3cm}|}
    \hline
     Operator & Single DON $100\%$ data & MODNO $100\%$ data &  MODNO $90\%$ data &  MODNO $80\%$ data  \\
    \hline
    Degree 2 & $2.52\%$ & $1.14\%$ & $1.3\%$ & $1.38\%$ \\
    \hline
    Degree 3 & $2.06\%$ & $2.03\%$ & $2.11\%$ & $2.25\%$ \\
    \hline
    Degree 4 & $2.6\%$ &  $2.54\%$& $2.79\%$ & $3.39\%$ \\
    \hline
    \end{tabular}
    \caption{The numerical results for the nonlinear porous media equations with different degrees. The second column, Single DON, is to train three individual DONs with identical structures with full data for each operator. }
    \label{table_porous_media}
\end{table}

\subsubsection{Experiment Two Results Analysis}
Observations from Table \ref{table_porous_media} highlight that MODNO surpasses the accuracy of training a single DON across three distinct operators. Even for equations of degree two and three, MODNO maintains superior accuracy compared to training a single DON. These findings are consistent with most of the outcomes from other numerical experiments, showing one potential benefit for the entire MOL society.

\subsection{Example Three}
\begin{figure}[H]
    \centering
    \includegraphics[scale = 0.4]{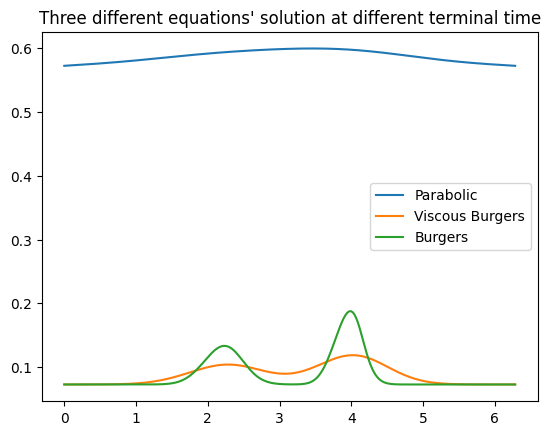}
    \caption{The terminal solutions of the parabolic equation (\ref{eqn_parabolic}), Viscous Burgers equation (\ref{eqn_viscous}) and Burger equation (\ref{eqn_burgers}) with the same initial conditions. Three equations' terminal simulation times are $0.5$, $1$ and $0.4$ respectively.  }
    \label{fig_parabolic_burger_v_terminal_solution}
\end{figure}

In this section, we will study applying a single network to learn the parabolic equation, 
\begin{align}
    u_t + u_{xx} = 1, x\in[0, 2\pi], t\in[0, 0.5], \label{eqn_parabolic}
\end{align} 
and the Viscous Burgers equation,
\begin{align}
    u_t + (\frac{u^2}{2})_x = \alpha u_{xx}, x\in[0, 2\pi], t\in[0, 1], \label{eqn_viscous}
\end{align}
where $\alpha = 0.1$, and the Burgers equation,
\begin{align}
    u_t + (\frac{u^2}{2})_x = 0, x\in[0, 2\pi], t\in[0, 0.4]. \label{eqn_burgers}
\end{align}
All equations are equipped with periodic boundary conditions. We study the mapping from the initial condition to the solutions. The initial conditions are generated as $w_1\mathcal{N}(\mu_1, \sigma_1^2)+w_1\mathcal{N}(\mu_1, \sigma_2^2)$ adjusted to be periodic with $w_1, w_2\sim \mathcal{U}(0, 0.5)$, $\mu_1, \mu_2\sim\mathcal{U}(2\pi, 4\pi)$, and  $\sigma_1, \sigma_2\sim\mathcal{U}(0.3, 1)$. For the demonstration, we present one realization of the terminal solution for three different types of equations in Figure \ref{fig_parabolic_burger_v_terminal_solution}. We can observe from the Figure that the three equations' solutions exhibit different behaviors, making the problem more challenging. 

We first introduce the relative errors for three distinct equations when forecasted utilizing the mean of all training target solutions. Specifically, the errors amount to $68.10\%$, $293.10\%$, and $281.34\%$ for the Parabolic, Viscous Burgers, and Burgers equations respectively.

We perform a sequence of numerical experiments to investigate the error fluctuation as we vary the number of samples used to train the shared branch nets. Additionally, we incorporate supplementary experiments where a single neural operator is trained to learn a single operator from the entire dataset (comprising all training samples), enabling a comparison with the multi-operator scenario. The outcomes are presented in Table \ref{table_exp_burgers}.
\begin{table}[H]
    \centering
    
    \begin{tabular}{|c|p{2.cm}|p{2.2cm}|p{2.2cm}|p{2.2cm}|p{2.2cm}|}
    \hline
     Operator & Single DON $100\%$ data & MODNO $100\%$ data &  MODNO $90\%$ data &  MODNO $80\%$ data &  MODNO $70\%$ data \\
    \hline
    Parabolic & $1.48\%$ & $0.22\%$ & $0.31\%$ & $0.5\%$ & $0.3\%$ \\
    \hline
    Viscous Burgers & $3.88\%$ & $2.96\%$ & $2.67\%$ & $3.24\%$ & $5.98\%$ \\
    \hline
    Burgers & $4.82\%$ & $5.34\%$ & $5.41\%$& $5.87\%$ & $7.53\%$ \\
    \hline
    \end{tabular}
    \caption{The numerical results for the Parabolic (\ref{eqn_parabolic}), Viscous Burgers equation (\ref{eqn_viscous}) and Burgers equation (\ref{eqn_burgers}). The second column, Single DON, is to train three individual DONs with identical structures with full data for each operator. Notably, when reducing MODNO data to $n\%$, MODNO has lower costs ($n\%$) when compared to three individual DONs. Please refer to Section \ref{sec_cost} for the cost evaluation details. }
    \label{table_exp_burgers}
\end{table}

\subsubsection{Experiment Three Results Analysis}
\label{sec_exp1_analysis}
From the results in Table \ref{table_exp_burgers}, it is evident that the errors of the MODNO for both the Burgers and Viscous Burgers equations exhibit a slight increase as the number of training samples decreases (improved efficiency). However, the prediction errors for the parabolic equation remain highly robust regardless of the number of training samples. Notably, our experiments involve applying a single neural operator (DON) with an identical structure (as the multi-operator MODNO) to learn a single operator. Each operator undergoes training using the same quantity of data as that used for training the full multi-operator model. As illustrated in Table \ref{table_exp_burgers}, even when reducing the amount of training data to $70\%$ of what is used for training a single neural operator, the multi-operator frameworks still outperform the single neural operators. 
Therefore, we can infer that the multi-operator MODNO yields satisfactory prediction outcomes while requiring fewer computational resources. This discovery holds greater significance when there is a scarcity of data available for a specific operator, as the data from other operators can aid in the learning process of the target operator.

\subsection{Example Four}
\begin{figure}[H]
    \centering
    \includegraphics[scale = 0.4]{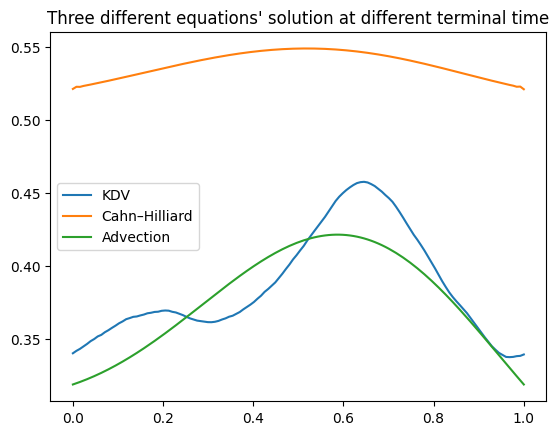}
    \caption{The terminal solutions of the KDV equation (\ref{eqn_kdv}), Cahn-Hilliard  (\ref{eqn_ch}) and Advection equation (\ref{eqn_adv}) with the same initial conditions. Three equations' terminal simulation times are $1$, $1$ and $0.1$ respectively. }
    \label{fig_kdv_ch_adv_terminal_sol}
\end{figure}

In this section, we will study mixing the Korteweg–De Vries (Kdv) equation, 
\begin{align}
    u_{t} +\delta^2 u_{xxx} +uu_x=0, x\in[0, 1], t\in[0,1],  \label{eqn_kdv}
\end{align}
where $\delta = 0.022$, Cahn-Hilliard (CH) equation,
\begin{align}
    u_{tt} +\epsilon^2 u_{xxxx} +6(uu_x)_x=0,  x\in[0, 1], t\in[0,1], \label{eqn_ch}
\end{align}
where $\epsilon = 0.01$, and Advection equation,
\begin{align}
    u_{t} + u_{x}=0,  x\in[0, 1], t\in[0,0.1]. \label{eqn_adv}
\end{align}
All equations are equipped with periodic boundary conditions. We study the mapping from the initial condition to the solutions at a later time. 
The initial conditions are generated as $w_1\mathcal{N}(\mu_1, \sigma_1^2)+w_1\mathcal{N}(\mu_1, \sigma_2^2)$ adjusted to be periodic with $w_1, w_2\sim \mathcal{U}(0, 0.5)$, $\mu_1\sim\mathcal{U}(0.1, 0.9)$, $\mu_2\sim\mathcal{U}(0.8, 1.5)$, and  $\sigma_1, \sigma_2\sim\mathcal{U}(0.1, 0.5)$. We present the terminal solutions with respect to the same initial condition for three different equations in Figure \ref{fig_kdv_ch_adv_terminal_sol}.

Similar to the previous examples, we first compute the mean prediction error, specifically, we use the mean of all target training solutions to predict all three different operators. The errors are $68.60\%$, $35.75\%$, and $76.68\%$ for KDV, Cahn-Hilliard and Advection equation respectively. We then present the results of the training and testing in Table \ref{table_kdv_results}.

\begin{table}[H]
    \centering
    \begin{tabular}{|c|p{2cm}|p{2.3cm}|p{2.3cm}|p{2.3cm}|p{2.3cm}|}
    \hline
     Operator & Single DON $100\%$ data & MODNO $100\%$ data &  MODNO $90\%$ data &  MODNO $80\%$ data & MODNO $70\%$ data \\
    \hline
    KDV    & $8.39\%$ & $7.62\%$ & $7.66\%$ & $7.53\%$ & $7.23\%$ \\
    \hline
    Cahn Hilliard & $0.28\%$ & $1.13\%$ & $1.49\%$ & $1.54\%$ & $2.21\%$ \\
    \hline
    Advection & $2.09\%$ & $4.5\%$ & $4.29\%$& $4.23\%$ & $3.96\%$ \\
    \hline
    \end{tabular}
    \caption{Examples Four Results. The second column, represents the training of three individual DONs with identical structures, each trained with full data for a single operator. For MODNO, we reduce the amount of data used in the shared structure and present the results in the last three columns. }
    \label{table_kdv_results}
\end{table}

\subsection{Experiment Four Results Analysis}
In Table \ref{table_kdv_results}, it is evident that MODNO outperforms single DON in predicting the KDV equation across all datasets with fewer training samples (more efficient). We notice a decrease in errors in both KDV and Advection prediction as we decrease the amount of data used to train the centralized shared structure. This partially confirms our conjecture that the centrally trained structure might overfit in some cases since it is trained by tripling the datasets using the same input.

\subsection{Experiment Five}
In this set of experiments,  we consider four equations one from each experiment in the previous sections, however, to even increase the level of difficulty, we choose different simulation terminal times.
Specifically, we will consider the nonlinear degree two porous media equation (\ref{eqn_porous_media}) with $t\in [0, 2]$, Sine-Gordon equation (\ref{eqn_sine}) with $t\in[0, 2]$, parabolic equation (\ref{eqn_parabolic}) with $t\in[0, 0.02]$ and Canh Hilliard equation (\ref{eqn_ch}) with $t\in[0, 0.5]$. 

\begin{figure}[H]
    \centering
    \includegraphics[scale = 0.4]{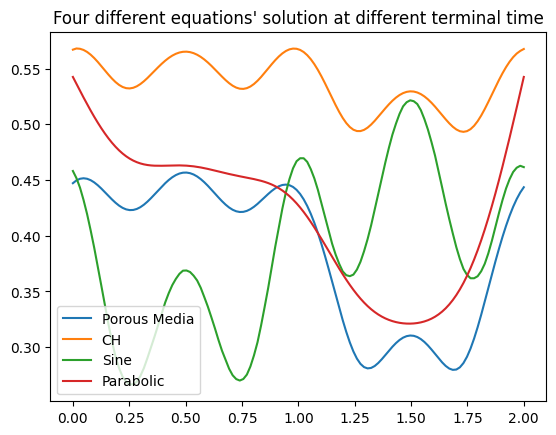}
    \caption{Demonstrations of one realization of terminal solutions corresponding to different equations. Porous media (\ref{eqn_porous_media}) at $t = 0.01$, Cahn Hilliard (CH) (\ref{eqn_ch}) at $t = 0.5$, Sine-Gordon (\ref{eqn_sine}) at $t = 2$ and Parabolic at $t = 0.02$.}
    \label{fig_four_terminal_sol}
\end{figure}

All equations are equipped with the periodic boundary conditions and we learn the mapping from the initial conditions to the solutions at a later time.
Specifically, we will test the extrapolation and predict the solution at terminal time which is not included in the training dataset. The initial conditions are generated the same as in Example 2. We present one realization of the solutions of four different equations at the terminal time in Figure \ref{fig_four_terminal_sol}. We summarize the numerical results in Table \ref{table_four_eqns}.

\begin{table}[H]
    \centering
    \begin{tabular}{|c|p{2.cm}|p{2.3cm}|p{2.3cm}|p{2.3cm}|}
    \hline
     Operator & Single DON $100\%$ data & MODNO $100\%$ data &  MODNO $90\%$ data &  MODNO $80\%$ data  \\
    \hline
    Porous Media & $0.98\%$ & $1.66\%$ & $1.17\%$ & $1.44\%$ \\
    \hline
    Cahn Hilliard & $3.24\%$ & $3.04\%$ & $3.01\%$ & $3.24\%$ \\
    \hline
    Sine Gordon & $5.62\%$ &  $5.07\%$& $5.27\%$ & $5.29\%$ \\
    \hline
     Parabolic & $2.12\%$ &  $2.61\%$& $2.29\%$ & $2.73\%$ \\
    \hline
    \end{tabular}
    \caption{The numerical results for Example five of mixing four equations. The second column, Single DON, is to train four individual DONs with identical structures with full data for each operator. Notably, when reducing MODNO data, MODNO has lower costs when compared to three individual DONs. Please refer to Section \ref{sec_cost} for the cost evaluation details.}
    \label{table_four_eqns}
\end{table}

\subsection{Experiment Five Results Analysis}
In these experiments, we combine four equations with distinct landscapes, as illustrated in Figure \ref{fig_four_terminal_sol}. Analysis from Table \ref{table_four_eqns} reveals that MODNO surpasses SOL in predicting Canh Hilliard equations and Sine-Gordon equations, even when only utilizing $80\%$ of the total samples for training the centralized structure. However, the predictions for Degree two Porous media equations and Parabolic equations fall short of expectations. Nonetheless, these predictions remain accurate ($1.17\%$ and $2.29\%$) and closely match those of single DONs, even when trained with only $90\%$ of the data used for single DONs.

\section{Conclusion}
\label{sec_conclusion}

In this work, we propose a distributed algorithm to enable Deep Operator Neural Network (DON) and other Chen-Chen-type neural operators to be a multi-operator learning (MOL) framework. The idea is to construct the basis for each operator separately using its own data while centralized learning the input function model reduction. The proposed framework can learn multiple operators simultaneously with less cost when compared to training a single neural operator to learn each operator separately. For many tasks, we even observe an improved accuracy. Compared to other MOL foundation models, the proposed methods have significantly less training cost and can be run even on personal devices.

However, there are some limitations to the methods. Unlike MOL foundation models, the current models have limited extrapolation capabilities to unseen operators. Addressing this limitation constitutes a potential area for future research. Specifically, exploring the feasibility of fine-tuning with or without data informed by physics will be pursued. Leveraging the success of multi-fidelity models, it is anticipated that the proposed method can be readily generalized to other unseen operators.

Additional future research directions include extending the framework to be discretization-invariant, allowing for different input functions to be discretized differently. This extension would offer alternatives to eliminate the ``same input function space'' assumptions made in the framework. 

Another potential avenue for future research involves the integration of physics-informed loss to tackle parametric PDEs with free parameters. This enhancement would enable MODNO to accommodate a broader range of operators generated based on sampling the free parameters of the base PDEs, thereby advancing it into a more sophisticated MOL framework.

\section{Acknoledgement}
The author expresses gratitude to Professor Hayden Schaeffer from the Department of Mathematics at UCLA for help. Additionally, the author extends appreciation to Ms. Jingmin Sun, a Ph.D. candidate from Carnegie Mellon University, for her generous help with numerical PDE solvers.

\bibliographystyle{abbrv}
\bibliography{references}
\end{document}